\documentclass[letterpaper, 10 pt, conference]{ieeeconf}  

\IEEEoverridecommandlockouts                              

\usepackage{graphics} 
\usepackage{xcolor}
\usepackage{pgfplots}
\usepackage{amsmath} 
\usepackage{amssymb}  
\usepackage{mathtools}
\usepackage{algpseudocode}
\usepackage{algorithm}
\usepackage{tikz}
\usepackage{subfigure}
\usepackage[acronym]{glossaries}
\usepackage{hyperref}
\usetikzlibrary{positioning}
\pgfplotsset{compat=1.18} 
\pgfplotsset{every tick label/.append style={font=\tiny}}
\definecolor{redvw}{HTML}{F36B9C}
\definecolor{bluevw}{HTML}{1E88E5}
\definecolor{greenvw}{HTML}{005648}
\newacronym{mpc}{MPC}{model predictive control}
\newacronym{mse}{MSE}{mean-squared error}
\newcommand\copyrighttext{%
  \footnotesize This work has been submitted to the IEEE for possible publication. Copyright may be transferred without notice, after which this version may no longer be accessible.}
\newcommand\copyrightnotice{%
\begin{tikzpicture}[remember picture,overlay]
\node[anchor=south,yshift=10pt] at (current page.south) {\fbox{\parbox{\dimexpr\textwidth-\fboxsep-\fboxrule\relax}{\copyrighttext}}};
\end{tikzpicture}%
}

\title{\LARGE \bf
Sampling-Based Model Predictive Control for Volumetric Ablation in Robotic Laser Surgery
}

\author{Vincent Y. Wang$^{* 1, 3}$, Ravi Prakash$^{1,3}$, Siobhan R. Oca$^{1, 3}$, Ethan J. LoCicero$^{1}$,\\ Patrick J. Codd$^{1, 2, 3}$, Leila J. Bridgeman$^{1, 3, 4}$
\thanks{$^{1}$Duke University, Thomas Lord Department of Mechanical Engineering and Materials Science.
}%
\thanks{$^{2}$Duke University, Department of Neurosurgery.}%
\thanks{$^{3}$Funded by the National Science Foundation Traineeship for the Advancement of Surgical Technologies.}%
\thanks{$^{4}$Funded by the National Science Foundation under Grant Number 2303158.}%
\thanks{* Corresponding author: Vincent Wang ({\tt\small vyw2@duke.edu})}%
}

\begin{document}

\maketitle
\copyrightnotice
\thispagestyle{empty}
\pagestyle{empty}

\begin{abstract}

Laser-based surgical ablation relies heavily on surgeon involvement, restricting precision to the limits of human error. The interaction between laser and tissue is governed by various laser parameters that control the laser irradiance on the tissue, including the laser power, distance, spot size, orientation, and exposure time. This complex interaction lends itself to robotic automation, allowing the surgeon to focus on high-level tasks, such as choosing the region and method of ablation, while the lower-level ablation plan can be handled autonomously. This paper describes a sampling-based \acrfull{mpc} scheme to plan ablation sequences for arbitrary tissue volumes. Using a steady-state point ablation model to simulate a single laser-tissue interaction, a random search technique explores the reachable state space while preserving sensitive tissue regions. The sampled \acrshort{mpc} strategy provides an ablation sequence that accounts for parameter uncertainty without violating constraints, such as avoiding critical nerve bundles or blood vessels.

\end{abstract}

\section{INTRODUCTION}

The use of energy-based laser scalpels in place of traditional mechanical scalpels represents a rapidly developing area of study, with existing applications in lithotripsy, neurosurgery, oncology, and cardiovascular procedures \cite{khalkhal2019evaluation}. However, most current laser scalpel procedures employ handheld operation, sometimes with robotic assistance \cite{basov2018robot}. With the absence of tactile feedback due to the non-contact nature of laser scalpels, surgeons must rely on training and estimation to remove tissues. Consequently, surgical outcomes depend on surgeon-to-surgeon skill, especially in robot-assisted, minimally-invasive procedures where common perceptive and haptic feedback mechanisms are absent. The precision required in surgical applications as well as the potential integration of various sensing modalities are thus well-suited to an autonomous robotic system.

To design automated laser ablation tools, it is helpful to have an accurate model for laser-tissue interactions. While previous works have studied these models, the interaction is a complex biophysical phenomenon \cite{lee2022end}. Lasers can have varying power densities, beam profiles, beam spot sizes, wavelengths, and orientations, while the tissue response depends on its refraction, scattering, absorption, and thermodynamic properties \cite{bordin2019laser}. A common steady-state model for tissue response to a single laser spot ablation employs a Gaussian model relying on the laser position and power, tissue density, and ablation enthalpy \cite{ross2018optimized}. Recently, this model has been extended to include the effect of laser angle on ablation \cite{ma2020characterization}, and online methods have been developed to obtain tissue parameters intraoperatively, enabling faster model-based planning and control methods \cite{fichera2016online,arnold2022identification,pardo2015learning,pacheco2024automatic}.

Despite previous work modeling laser-tissue interactions, a gap remains between real target tissue and the test models and settings used in the laboratory. In reality, tissues have heterogeneous material properties and complex geometries, and as such, a sequence of multiple cuts must be planned to resect a volumetric region. Previous volumetric ablation studies have used a raster-based movement pattern that sweeps across the tissue surface \cite{ross2018optimized} or 2D layered packing algorithms \cite{kahrs2010planning}. However, these methods are prone to error compounded through layers, or in tumors that may have heterogeneities. They also do not utilize angled cuts, restricting the output space of removable tumor shapes.

Beyond the lack of generalizability, the lack of any feedback limits the system's ability to respond to model uncertainty and compounds the risk of ablating critical anatomical structures. \Acrfull{mpc} represents a common solution from industrial control. It has precedence in the medical space for applications such as a linear \acrshort{mpc} formulation for tissue temperature control in targeted ultrasonic heating therapy for cancer treatment via thermal ablation \cite{hensley2015model}. However, classical \acrshort{mpc} often requires linearized approximations \cite{darby2012mpc}. In systems with strong nonlinearities, this degrades controller performance. This can be particularly challenging in biological tissues, where material properties and mechanics vary dramatically under different conditions, often requiring highly nonlinear models \cite{freutel2014finite}.

This paper presents a graph-based planning and control algorithm to achieve laser ablation on arbitrary tissue volumes. By planning multiple single ablations using a Gaussian steady-state ablation model and a modified sampling-based \acrshort{mpc} formulation commonly used in robotics and planning problems \cite{dunlap2008sampling,reese2016graph}, the proposed method is capable of resecting to a desired boundary profile and is adaptable to a wide range of desired ablation geometry, as well as having the ability to remain constraint-aware and preserve prohibited tissue regions. A comparison is also performed to a basic nonlinear optimization method to demonstrate the benefits and drawbacks of the sampling-based method.

\section{METHODS}

\subsection{Single-Point Ablation Model}
\begin{figure}[b]
\centering
\begin{tikzpicture}
  \begin{axis}[ticks=none,axis x line=none,axis y line=none,
      xticklabel=\empty,
      yticklabel=\empty,
    ]
    \draw[->, dashed, color=redvw] (-0.3965,-0.4587) -- (0.4458, 0.5413) node[above]{Test};
    \draw[-, dashed, color=greenvw] (0.1518,-0.2636) -- (-0.202, -0.24);
    \draw[-, dashed, color=bluevw] (0.1518,-0.2636) -- (0.3738, 0);
    \draw[-, dashed, color=black] (-1, 0) -- (1, 0);
    \filldraw[black] (0.1518,-0.2636) circle (2pt) node[anchor=north west, outer sep=1]{$\vec{p}_+$};
    \filldraw[black] (0.3738, 0) circle (2pt) node[anchor=north west, outer sep=1]{$\vec{p}$};
    \filldraw[black] (0, 0) circle (2pt) node[anchor=north west, outer sep=1]{$x_L$};
    \filldraw[black] (-0.1, -0.2) circle (0pt) node[anchor=north west, outer sep=0]{\textcolor{greenvw}{$d$}};
    \filldraw[black] (-0.03, -0.08) circle (0pt) node[anchor=north west, outer sep=1]{\textcolor{bluevw}{$\Delta p$}};
    \addplot [color=black]
      table[row sep=crcr]{%
   -1.0000         0\\
   -0.9798         0\\
   -0.9596         0\\
   -0.9394         0\\
   -0.9192         0\\
   -0.8990         0\\
   -0.8788         0\\
   -0.8586         0\\
   -0.8384         0\\
   -0.8182         0\\
   -0.7980         0\\
   -0.7778         0\\
   -0.7630   -0.0064\\
   -0.7500   -0.0150\\
   -0.7375   -0.0241\\
   -0.7256   -0.0339\\
   -0.7141   -0.0444\\
   -0.7032   -0.0554\\
   -0.6929   -0.0671\\
   -0.6830   -0.0794\\
   -0.6737   -0.0923\\
   -0.6649   -0.1058\\
   -0.6565   -0.1199\\
   -0.6486   -0.1345\\
   -0.6411   -0.1495\\
   -0.6339   -0.1650\\
   -0.6271   -0.1809\\
   -0.6205   -0.1971\\
   -0.6142   -0.2135\\
   -0.6080   -0.2301\\
   -0.6018   -0.2468\\
   -0.5957   -0.2636\\
   -0.5895   -0.2802\\
   -0.5832   -0.2966\\
   -0.5766   -0.3128\\
   -0.5697   -0.3286\\
   -0.5624   -0.3439\\
   -0.5546   -0.3587\\
   -0.5463   -0.3728\\
   -0.5373   -0.3861\\
   -0.5276   -0.3985\\
   -0.5170   -0.4100\\
   -0.5056   -0.4204\\
   -0.4933   -0.4297\\
   -0.4799   -0.4378\\
   -0.4655   -0.4447\\
   -0.4500   -0.4503\\
   -0.4333   -0.4545\\
   -0.4155   -0.4573\\
   -0.3965   -0.4587\\
   -0.3763   -0.4587\\
   -0.3549   -0.4573\\
   -0.3323   -0.4545\\
   -0.3086   -0.4503\\
   -0.2837   -0.4447\\
   -0.2577   -0.4378\\
   -0.2306   -0.4297\\
   -0.2026   -0.4204\\
   -0.1736   -0.4100\\
   -0.1437   -0.3985\\
   -0.1131   -0.3861\\
   -0.0817   -0.3728\\
   -0.0496   -0.3587\\
   -0.0170   -0.3439\\
    0.0161   -0.3286\\
    0.0497   -0.3128\\
    0.0835   -0.2966\\
    0.1175   -0.2802\\
    0.1518   -0.2636\\
    0.1860   -0.2468\\
    0.2203   -0.2301\\
    0.2545   -0.2135\\
    0.2885   -0.1971\\
    0.3224   -0.1809\\
    0.3560   -0.1650\\
    0.3892   -0.1495\\
    0.4221   -0.1345\\
    0.4546   -0.1199\\
    0.4866   -0.1058\\
    0.5182   -0.0923\\
    0.5493   -0.0794\\
    0.5798   -0.0671\\
    0.6099   -0.0554\\
    0.6394   -0.0444\\
    0.6968   -0.0241\\
    0.7248   -0.0150\\
    0.7522   -0.0064\\
    0.7778         0\\
    0.7980         0\\
    0.8182         0\\
    0.8384         0\\
    0.8586         0\\
    0.8788         0\\
    0.8990         0\\
    0.9192         0\\
    0.9394         0\\
    0.9596         0\\
    0.9798         0\\
    1.0000         0\\
    };
    \end{axis}
    \filldraw[black] (2, 6.3) circle (0pt) node[anchor=north west, outer sep=1]{Laser Center Axis};
\end{tikzpicture}
\caption{The effect of a single-point ablation according to the model \eqref{eq:singleablation}. A given point, $\vec{p}$, is displaced along the direction of the laser axis by a distance, $\Delta p$, as a Gaussian function of the normal distance to the axis, $d$. The laser position is marked as $x_L$.} \label{fig: singleablation-diagram}
\end{figure}
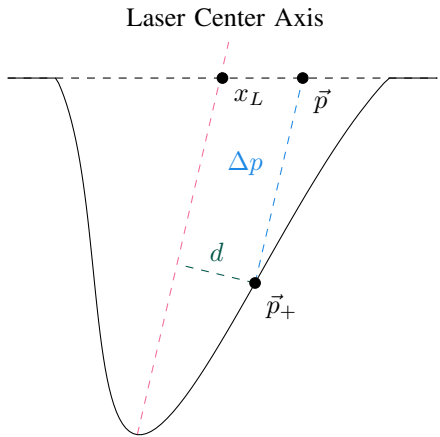

Here, two different tissue ablation planning methods are proposed based on the discrete time, point-ablation model, 
\begin{align} \label{eq:singleablation}
    \Delta p &= \frac{1}{\beta}\mathrm{max}\left(E \Delta t e^{-2\left(\frac{d^2}{w^2}\right)}-\phi\right),
\end{align}
which assumes a Gaussian beam profile centered at a point, $\vec{p}$, on the tissue surface \cite{ross2018optimized}. After an ablation lasting $\Delta t$, the point displaces parallel to the laser axis by $\Delta p$, which is a function of the density of tissue multiplied by the ablation enthalpy of the tissue, $\mathrm{\beta}$, the minimum energy threshold required to begin the ablation process, $\phi$, the spot size of the laser, $w$, the laser power, $E$, and the orthogonal distance between the point $\vec{p}$ and the laser axis, $d$. The orthogonal distance $d$ also depends on the laser's incident point on the $x$-axis, $x_L$, and the laser angle from vertical, $\theta_L$. The effect of a laser ablation at a single point is shown in \autoref{fig: singleablation-diagram}.

\subsection{Nonlinear Optimization Algorithm}
The first planning method poses a nonconvex optimization problem to obtain the optimal laser power at a set of discrete spatial points, $\vec{x}=[x_1, x_2 \dots x_n]$. At each $x_i$, this method selects the optimal power, $E_i$, to minimize the \acrfull{mse} between the desired and actual boundary profiles in the $z$-direction.  
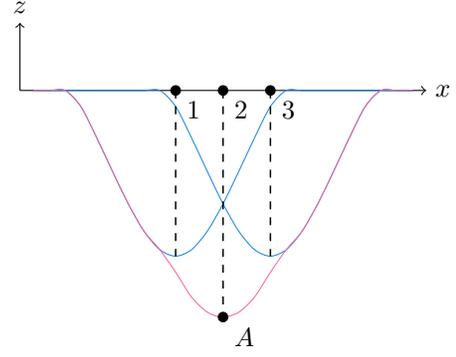
\begin{figure}
\centering
\vspace*{0.1cm}
\begin{tikzpicture}[scale=0.9]
  \draw[->] (-3, 0) -- (3, 0) node[right] {$x$};
  \draw[->] (-3, 0) -- (-3, 1) node[above] {$z$};
  \draw[scale=0.7, domain=-4:4, smooth, variable=\x, bluevw] plot ({\x}, {-max(0, 5*exp(-2*(\x-1)*(\x-1)/8)-1.5)});
  \draw[scale=0.7, domain=-4:4, smooth, variable=\x, bluevw] plot ({\x}, {-max(0, 5*exp(-2*(\x+1)*(\x+1)/8)-1.5)});
  \draw[scale=0.7, domain=-4:4, smooth, variable=\x, redvw] plot ({\x}, {-max(0, 5*exp(-2*(\x+1)*(\x+1)/8)-1.5)-max(0, 5*exp(-2*(\x-1)*(\x-1)/8)-1.5)});
  \filldraw[black] (0, 0) circle (2pt) node[anchor=north west, outer sep=1]{$2$};
  \filldraw[black] (0.7, 0) circle (2pt) node[anchor=north west, outer sep=1]{$3$};
  \filldraw[black] (-0.7, 0) circle (2pt) node[anchor=north west, outer sep=1]{$1$};
  \filldraw[black] (0, -3.35) circle (2pt) node[anchor=north west, outer sep=1]{$A$};
  \draw[-, line width=0.2mm, dashed] (0.7, 0) -- (0.7, -2.45);
  \draw[-, line width=0.2mm, dashed] (-0.7, 0) -- (-0.7, -2.45);
  \draw[-, line width=0.2mm, dashed] (0, 0) -- (0, -3.35);
\end{tikzpicture}
\caption{Linear superposition of individual ablations with zero angular components. The net ablation (red) is a combination of two single ablations (blue). Using the notation defined in \eqref{eq:superposition}, the ablation in this figure illustrates that the distance between point $2$ and point $A$ is $\Delta p_{T2}=\Delta p_{12}+\Delta p_{32}$.} \label{fig: superposition-diagram}
\end{figure}
\begin{figure}
\centering
\input{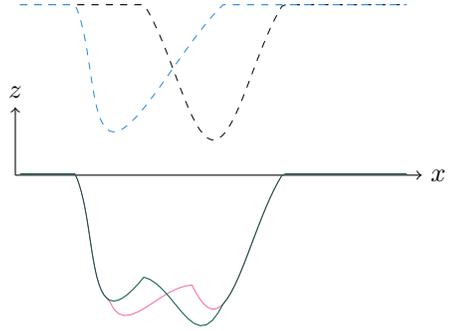}
\caption{Two laser profiles are shown: profile 1 in black with laser settings $\left[x_L=0, \theta_L=0, E_L=5\right]$ and profile 2 in blue with laser settings $\left[x_L=-0.25, \theta_L=0.3491, E_L=5\right]$. The net ablation that results from applying cuts in the order of profile 1-2 (red) differs from the net ablation that results from applying cuts in the order of profile 2-1 (green).} \label{fig:angularsup}
\end{figure}
Two additional constraints are imposed to render a tractable optimization problem. First, the laser angle is fixed to zero (i.e. pointing vertically downwards) to exploit linear superposition, as the effect of multiple cuts will add linearly in the $z$-direction (\autoref{fig: superposition-diagram}, proof in Appendix A). Superposition does not hold for cuts at different angles (\autoref{fig:angularsup}). Second, only one cut is made per location $x_i$. These assumptions sacrifice input options, but reduce the optimization problem to finding the vector of power settings, $\vec{E}\in \mathbb{R}^n$, for each point, $\vec{x}$. 

The total cut depth can be found by summing the contributions of each cut as
\begin{align} \label{eq:superposition}
    \Delta p_{Tj} &= \sum_{i=1}^{n} \Delta p_{ij},
\end{align}
where $\Delta p_{Tj}$ is the total cut depth at position $x_j$, and $\Delta p_{ij}$ is the depth contribution at position $x_j$ from the laser firing at position $x_i$, as depicted in \autoref{fig: superposition-diagram}. 
Using \eqref{eq:singleablation}, $\Delta p_{Tj}$ at every point $x_j$ can be expanded to
\begin{align*}
    \Delta \vec{p} = 
    \begin{bmatrix}
        \Delta p_{T1}\\
        \vdots\\
        \Delta p_{Tn}
    \end{bmatrix} &= \frac{1}{\beta} \mathrm{max}\left(0, 
    \mathbf{E} \odot \mathbf{P}
    -
    \phi\mathbf{1}_{n\times n}
    \right)
    \begin{bmatrix}
        1\\
        \vdots\\
        1
    \end{bmatrix}\\
    P_{i:j} &\coloneqq \Delta t\, \mathrm{exp}\left(-2\left(\frac{\left(x_i^2-x_j^2\right)}{w^2}\right)\right),
\end{align*}
where $\odot$ is defined as the Hadamard product, $\mathbf{E}$ is defined as the $n \times n$ matrix resulting from the length-$n$ row vector $\vec{E}$ stacked vertically $n$ times, and $\mathbf{P}$ is the constant $n \times n$ matrix where row $i$, column $j$ contains the constant $P_{i:j}$. This formulation gives the final problem $\min_{\vec{E}\geq0}\,\| \Delta \vec{p}-\vec{p}_d\|_2 \ \mathrm{s.t.}\ \Delta \vec{p}-\vec{p}_c\leq0$,
which finds a power input, $E_i$, at each point that produces a local cost minimum given an objective boundary, $\vec{p}_d$, and constraint boundary, $\vec{p}_c$. 
\subsection{Graph Search Algorithm}
Though the nonlinear optimization approach can be solved across the entire time domain in a single iteration, prohibiting angled and repeat cuts constrains the laser operational range. Consequentially, it may perform poorly on volumes with complex or angled geometry, such as the sample profile in \autoref{fig: singleablation-diagram}. To expand the input space, a heuristic planning method is proposed using a graph search problem similar to other methods found in robotics literature \cite{dunlap2008sampling}. 

\begin{figure}
\centering
\vspace*{0.2cm}
\begin{tikzpicture}[
graphNode/.style={circle, draw=gray!60, fill=gray!5, very thick, minimum size = 8mm},
emphasis1/.style={circle, draw=bluevw!60, fill=gray!0, dotted, very thick, minimum size = 12mm},
emphasis3/.style={circle, draw=greenvw!90, fill=gray!0, dotted, very thick, minimum size = 8mm},
]
\node[graphNode]at (0, 0)(root){$S_0$};
\node[graphNode]at (-0.8, -1.5)(n2){$S_1$};
\node[emphasis1]at (0.8, -1.5)(e1){};
\node[graphNode]at (0.8, -1.5)(n1){$S_2$};
\node[graphNode]at (0, -3)(n3){$S_3$};
\node[emphasis3]at (1.6, -3)(e3){};

\draw[->, line width=0.2mm] (root) -- (n1);
\draw[->, line width=0.2mm] (root) -- (n2);
\draw[->, line width=0.2mm] (n1) -- (n3);
\draw[->, line width=0.5mm, dotted, draw=redvw!60] (e1) -- (e3);
\filldraw[black] (1.2, -2.25) circle (0pt) node[anchor=west, outer sep=10]{\textcolor{redvw!60}{$[x_L, \theta_L, E_L]$}};
\end{tikzpicture}
\caption{A visual representation of \autoref{alg1}. A random node is selected (blue), after which a random input is applied (red), leading to a new node (green) appended to the graph.} \label{fig: graph-diagram}
\end{figure}
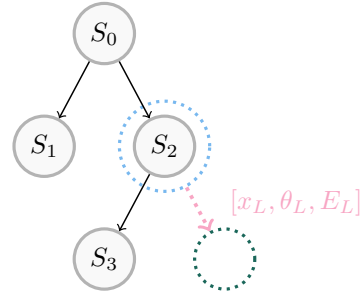

\begin{algorithm}
\caption{Laser parameter graph search}
\label{alg1}
\begin{algorithmic}
\State Graph $\gets$ initialize tree with initial state
\Repeat
    \State currentNode $\gets$ randomly sample a node from \texttt{Graph}
    \State input $\gets$ randomly sample from the input space, ($\left[X_L, \Theta_L, P_L\right]$)
    \State nextNode $\gets$ simulate ablation with (\texttt{currentNode}, \texttt{input})
    \If {\texttt{nextNode} does not violate constraints}
        \State Graph.Nodes $\gets$ add \texttt{nextNode}
    \EndIf
\Until{number of nodes in \texttt{Graph} exceeds $k_F$}
\State \Return node with the lowest objective cost within \texttt{Graph}
\end{algorithmic}
\end{algorithm}
\autoref{alg1} constructs a tree whose nodes contain a system state $S$, an $n \times D$ array of the $n$ points in a $D$-dimensional point cloud denoting the air-tissue boundary (\autoref{fig: graph-diagram}). Each edge contains a tuple of inputs, $\left[x_L,\theta_L,E_L\right]$, representing a single ablation location, angle, and power. The tree is initialized with a single node containing the initial state of the tissue. To expand the tree, the tree is sampled for a random node, representing a system state, and the input space is sampled for a random input vector to represent an edge. Each input is sampled from the set of allowable inputs for laser position, angle, and power, $x_L \in X_L,\theta_L \in \Theta_L,E_L \in P_L$, then substituted into the point ablation model alongside the randomly sampled node \eqref{eq:singleablation} to produce a new state. If the new state violates constraints, the node is not added to the tree; otherwise, it is added as a new node. For each state, an objective cost is also computed. After searching for $k_F$ nodes (akin to an $n$-step horizon in \acrshort{mpc}) or after a target cost is reached, the algorithm returns the node with the lowest cost and the corresponding inputs. 

Compared to the previous method, this algorithm expands the one-dimensional input space ($E_L$) in two dimensions ($X_L$ and $\Theta_L$), expanding the set of reachable states. Running the algorithm repeatedly produces subsequent inputs, and is completed once the cost improvement over a single run of the algorithm falls below a threshold, $\epsilon_c$.

To reduce the runtime of \autoref{alg1}, a weighting system is applied to the random sampling steps. Three possible weighting heuristics are described below. The angular input space $\Theta_L$ is sampled uniformly, so it is not mentioned below. 

\subsubsection{Nodal Sampling}
In the absence of weighted sampling, the algorithm will sample uniformly from all nodes in the tree. This provides a uniform search of the state space, but yields slow objective cost improvements due to the tendency to search from very shallow nodes. Instead, the selection of nodes is weighted towards lower cost nodes, where each node is assigned a weight of 
\begin{align*}
    w_i = \left(\mathrm{max}\left(\vec{C}^*\right)-C^*_i\right)^a+\epsilon_n,
\end{align*}
where $\vec{C}^* = \left[C^*_1, C^*_2 \dots C^*_k\right]$ and $C^*_i$ is a modified version of the original error objective cost defined as
\begin{align*}
    C^*_i &= \|\mathrm{min}\left(0, \Delta \vec{z}\,\right)\|^2_2+\lambda\|\mathrm{max}\left(0, \Delta \vec{z}\,\right)\|^2_2,\\
    \Delta z_i &= z_d(x_i)-z(x_i),
\end{align*}
where $z(x_i)$ and $z_d(x_i)$ are defined as the true and desired boundary $z$ coordinates for a given point respectively, and $\lambda\geq1$ is a tuneable parameter. This cost function is designed to distinguish overcut points (i.e. ablated past the desired boundary) and undercut points (i.e. not yet ablated to the target boundary). The $\lambda$ parameter discourages overcutting by assigning a higher penalty to overcut tissue, as any excess tissue removal cannot be ``undone'' by any feasible control inputs.
A higher probability is assigned to low-cost nodes, $C^*_i$, encouraging exploration of routes with existing low cost. The addition of a small $\epsilon_n$ term ensures nodes will have a positive nonzero weight for random sampling. The exponent $a$ is chosen experimentally to balance the promotion of exploration of low-cost nodes with permitting exploration of other paths to avoid being trapped in local minimum routes. 

\subsubsection{Laser Position Sampling}
The space of allowable laser positions $X_L$ is the set of discrete locations of each point in the point cloud. Each point $p_i$ in the point cloud has weight
\begin{align*}
    w_i &= \mathrm{cost}\left(p_i\right)+\epsilon_L,
\end{align*}
where $\mathrm{cost}\left(\cdot\right)$ is the objective cost function. This weighting promotes positioning the laser over areas of high objective cost, as they are more likely to have both a higher number of legal cuts available, as well as a higher potential for cost reduction. The $\epsilon_L$ term ensures positive, nonzero weights.

\subsubsection{Laser Power Sampling}
Given a discretized set of input power values to sample from, $\vec{E}_I$, the weight, $w_i$, assigned to a specific input power, $E_i$, is
\begin{align*}
    w_i &= e^{b\left(\mathrm{max}\left(\vec{E}_I\right)-\left|E_i-E_{p}\right|\right)}.
\end{align*}
For a chosen laser position, $x_L$, from the previous sampling step, $E_p$ is defined as the predicted power required to cut a distance of $\left|z_d(x_L)-z(x_L)\right|$, where $z_d$ is the $z$-value of the objective boundary at $x_L$. In other words, $E_p$ is the power required to ablate the point at which the laser is currently centered to the objective boundary in one single cut. From the single-point ablation model \eqref{eq:singleablation}, $E_p$ can be derived as
\begin{align*}
    E_p &= \frac{\beta\left|z_d(x_L)-z(x_L)\right|+\phi}{\Delta t}.
\end{align*}
An additional tuning parameter $b$ is introduced into the exponent as to control the strength of the weighting scheme.
\subsection{Feedback Control Loop}
Both models can run in either a feedforward or feedback mode. To run in a feedforward mode, the algorithms are only run once, and the resulting input sequence is implemented without correction. To incorporate feedback, either model can act as the controller in a simple feedback control system. After each algorithm is run, the first input in the resultant input sequence is simulated as a cut using the single-point ablation model, after which the algorithm is re-run using sensed data about the resultant cut as the new initial state.

\vspace*{0.5cm}
\begin{figure*}[!htb]
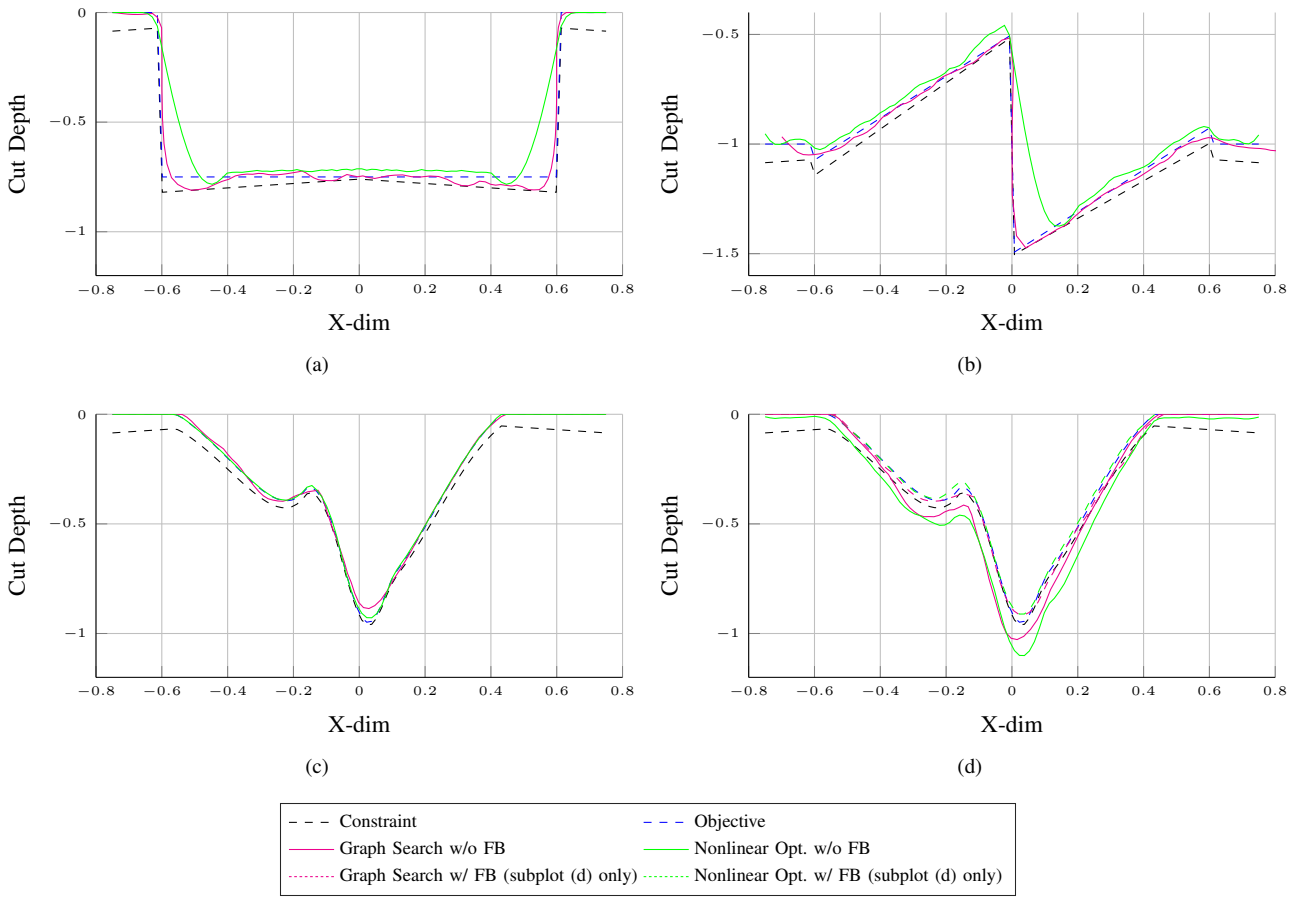
\centering
    \subfigure[]{\centering\label{fig:results_a}\input{flat}}
    \subfigure[]{\centering\label{fig:results_b}\input{sawtooth}}
    \subfigure[]{\centering\label{fig:results_c}\input{two-cut}}
    \subfigure[]{\centering\label{fig:results_d}\input{two-cut_uncertainty}}
    \subfigure{\centering\label{fig:results_legend}
%
%
\begin{tikzpicture}[font=\scriptsize] 

\begin{axis}[%
width=6.5cm,
height=3cm,
at={(0cm,0cm)},
hide axis, 
xmin=-0.8,
xmax=0.8,
ymin=-1.6,
ymax=-0.4,
legend style={at={(0.5,-0.3)}, anchor=north, legend columns=2, legend cell align=left, align=left, draw=white!15!black}
]       
\addlegendimage{black,dash pattern=on 3pt off 3pt on 3pt off 3pt}
\addlegendentry{Constraint};
\addlegendimage{blue,dash pattern=on 3pt off 3pt on 3pt off 3pt}
\addlegendentry{Objective};
\addlegendimage{magenta}
\addlegendentry{Graph Search w/o FB};
\addlegendimage{green}
\addlegendentry{Nonlinear Opt. w/o FB};
\addlegendimage{magenta,dash pattern=on 1pt off 1pt on 1pt off 1pt}
\addlegendentry{Graph Search w/ FB (subplot (d) only)};
\addlegendimage{green,dash pattern=on 1pt off 1pt on 1pt off 1pt}
\addlegendentry{Nonlinear Opt. w/ FB (subplot (d) only)};
\end{axis}

\end{tikzpicture}%
}
    \caption{Results of numerical simulations. \autoref{fig:results_a}, \autoref{fig:results_b}, \autoref{fig:results_c} show a comparison between the nonlinear optimization algorithm and the graph search algorithm with nominal system values for a square well, sawtooth, and two-cut boundary respectively. Note that since the system is nominal and has no error, only a feedforward method is presented. \autoref{fig:results_d} displays a repeat of the two-cut experiment, but includes a 5\% error between the nominal system values provided to the controller and the real parameters simulated by the plant. FB denotes ``feedback''.} \label{fig:results}
    \vspace*{-0.4cm}
\end{figure*}
\section{NUMERICAL EXPERIMENTS}
\subsection{Planning Algorithm Comparison}
The two algorithms were tested in open-loop on three 2D objectives with a 100-point ($n=100$) resolution using nominal parameter values. The test was performed on a square well, a sawtooth pattern, and a two-cut objective boundary created by simulating two manually-selected laser inputs. The constraint boundary was defined as $z_c(x) = z_d(x)-a|x|-b$ with constants $a$ and $b$ to simulate variable constraint depth. The graph search ran with $k_F\approx10^5$ per step. \autoref{fig:results} presents the results and \autoref{tb:nommse} presents metrics.
\begin{table}[b]
\centering
\caption{Open Loop Performance for Volumetric Ablation Algorithms}
\label{tb:nommse}
\begin{tabular}{|l||l|l|l|l|}
\hline
& \multicolumn{2}{c|}{Graph Search} & \multicolumn{2}{c|}{Nonlinear Opt} \\
\hline
 & MSE & Time (min) & MSE & Time (min) \\
\hline\hline
Square Well & $\mathbf{1.49\mathrm{\textbf{E}}{-2}}$ & $13.7$ & $1.72\mathrm{E}{-2}$ & $<1$\\
\hline
Sawtooth & $\mathbf{1.14\mathrm{E}{-2}}$ & $18.6$ & $1.84\mathrm{\textbf{E}}{-2}$ & $<1$\\
\hline
Two-Cut & $24.5\mathrm{E}{-5}$ & $15.2$ & $\mathbf{2.63\mathrm{\textbf{E}}{-5}}$ & $<1$\\
\hline
\end{tabular}
\vspace*{-0.7\baselineskip}
\end{table}
\subsection{Feedback Control for Uncertainty Compensation}
To study the impact of feedback in mitigating model uncertainty, the experiments were repeated on the two-cut boundary, with the three inherent tissue parameters used in the ablation simulator (density and ablation enthalpy, both part of $\beta$, and the energy threshold, $\phi$) decreased by 5\% from the nominal value given to the controller. The results are presented in \autoref{fig:results_d}, with metrics and constraint violations given in \autoref{tb:errormse}.

\begin{table}[b]
\centering
\caption{Ablation Algorithm Performance Under Uncertainty}
\label{tb:errormse}
\begin{tabular}{|l||l|l|l|l|}
\hline
& \multicolumn{2}{c|}{Graph Search} & \multicolumn{2}{c|}{Nonlinear Opt} \\
\hline
 & FFwd & Fdbk & FFwd & Fdbk \\
\hline\hline
MSE & $52.8\mathrm{E}{-4}$ & $4.06\mathrm{E}{-4}$ & $88.2\mathrm{E}{-4}$ & $1.54\mathrm{E}{-4}$\\
\hline
\% Violation & 42\% & 14\% & 58\% & 0\%\\
\hline
Runtime (min) & 14 & 62 & $<1$ & 60\\
\hline
\end{tabular}
\vspace*{-0.7\baselineskip}
\end{table}

\subsection{3D Simulations on Brain Tumor Volume Data}
\begin{figure*}[!htb]\centering
    \subfigure[]{\centering\label{fig:3dsetup}\includegraphics[width=0.4\textwidth]{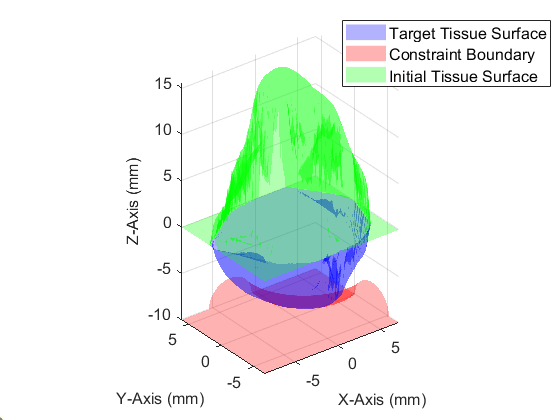}}
    \subfigure[]{\centering\label{fig:3dfinal}\includegraphics[width=0.4\textwidth]{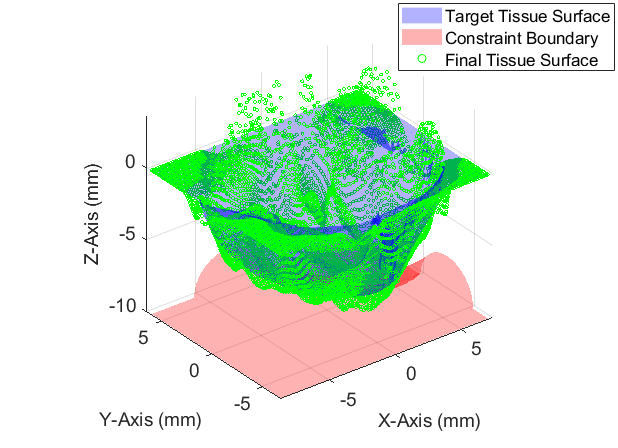}}
    \\\vspace*{-0.35cm}
    \subfigure[]{\centering\label{fig:3derror}\includegraphics[width=0.4\textwidth]{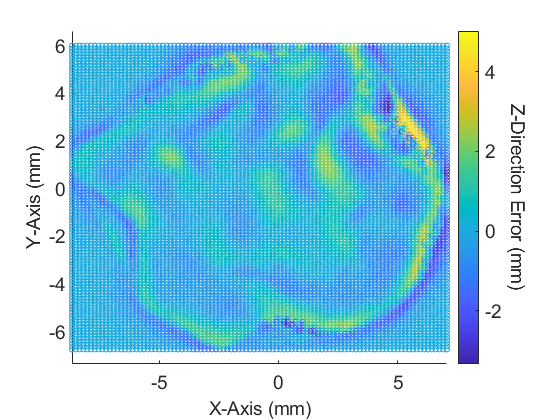}}
    \subfigure[]{\centering\label{fig:3dhistogram}\includegraphics[width=0.4\textwidth]{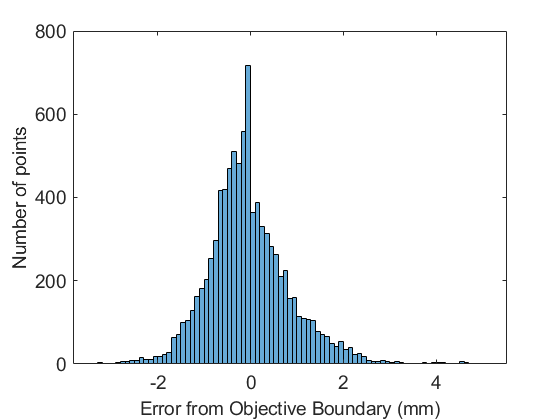}}

    \caption{Results from a 3D volumetric simulation of a brain tumor ablation using the graph-sampling method. (a) The initial tumor boundary, objective, and constraint surfaces. (b) The final ablated crater post-algorithm. (c) A top-down view of error measured as the distance between the final tissue surface and the objective surface at each point in the $z$-direction. Negative values indicate overcuts, positive values indicate undercuts. (d) A histogram of error across 9009 total points (with 991 trivial noncut points around the flat border region removed to include only points affected by the ablation sequence).} \label{fig:3dresults}
    \vspace*{-0.4cm}
\end{figure*}

The graph-search algorithm was also tested on a 3D sample of real brain tumor data extracted from an MRI scan. The nonlinear-optimization algorithm was attempted as well; however, the space requirement scales exponentially with the number of dimensions, and the algorithm terminated due to memory limits. MRI data taken from the 2017 Multimodal Brain Tumor Segmentation dataset \cite{menze2014multimodal,bakas2017advancing,bakas2018identifying,bakas2017segmentationgbm,bakas2017segmentationlgg} was segmented using the 3D Slicer software \cite{3dslicer,fedorov20123d}. The tumor was embedded into a flat plane to define the objective boundary, and a constraint boundary was generated using a quarter-torus to simulate a critical blood vessel near the tumor (\autoref{fig:3dsetup}). The tissue surface was represented with a $100 \times 100$ point cloud.
The modified three-dimensional algorithm includes two new dimensions in the input space, $[X_L, Y_L, \Theta_{xL}, \Theta_{yL}, P_L]$, and uses a two-dimensional distance equation to find $d$ in \eqref{eq:singleablation}. Total runtime for the planning algorithm was $\sim3$ hours, and the results are shown in \autoref{fig:3dresults}, with evaluation parameters in \autoref{tb:3dmetrics}. 
\begin{table}[b]
\centering
\caption{Summary Statistics for 3D Volumetric Ablation Simulation}
\label{tb:3dmetrics}
\begin{tabular}{|l||l|}
\hline
MSE & 0.712\\
\hline
Original Tumor Volume (OTV) (mm$^3$) & 1830.38\\
\hline
Removed Healthy Tissue Volume (mm$^3$) & 63.13 (3.45\% OTV)\\
\hline
Remaining Tumor Volume (mm$^3$) & 56.71 (3.10\% OTV)\\
\hline
Number of Constraint-Violating Points & 0\\
\hline
\end{tabular}
\vspace*{-0.7\baselineskip}
\end{table}

\section{DISCUSSION}
In 2D, the nonlinear optimization method had lower MSE only on the two-cut boundary. It uses linear superposition to formulate the problem, requiring no angular or repeat ablations, meaning it performs sub-optimally with steep walls that may benefit from an angled cut or regions where follow-up cuts are required. This is visible in the square well and sawtooth of \autoref{fig:results}, where there are large uncut regions near the sharp corners. Conversely, the graph search method frequently overcuts past the objective, visible in \autoref{fig:results_a}. Initially, the algorithm favors high-power cuts that remove a large volume of tumor tissue. These cuts may overcut past the objective boundary slightly; however, since a large volume of tumor tissue is removed, the cuts still greatly reduce the objective cost. The small overcut is then permanent, as an overcut region cannot be ``uncut'' in the future. This paper modified the cost function during exploration to preferentially penalize overcuts, discouraging input sequences with overcuts. Another solution is to tighten the constraint boundary to create a new ``pseudo''-constraint boundary that prevents overcuts. As a test, the two-cut nominal simulation was repeated equating the objective and constraint boundaries. Tightening this constraint reduced the two-cut graph search \acrshort{mse} (\autoref{tb:nommse}) by over $50\%$; from $24.5\mathrm{E}{-5}$ to $11.4\mathrm{E}{-5}$. Future work will seek a systematic way to set pseudo-constraint boundaries.

Tumors are often continuous, amorphous, and irregular volumes, and the graph search method will provide a smarter ablation sequence. If the constraint boundary is lax (leading to more overcutting) or the tumor geometry has an amenable topography, the nonlinear optimization algorithm may be adequate and quicker. The methods can also be used in tandem---the nonlinear optimization algorithm runs quickly but has reduced input options and a large memory requirement, so it may be suited for an initial, low-resolution tumor debulking. The graph method can subsequently precisely remove residual tumor with more complex cut requirements.

The uncertainty simulation displays the compounding effect of small model inconsistencies across many cuts, having constraint violations near $50\%$ without feedback. The feedback controller greatly reduced constraint violations, as seen in \autoref{tb:errormse}, but could not preclude single cuts that immediately violate constraints, leading to the minute violations seen in \autoref{fig:results_d}. Future work may incorporate robust, constrained controllers, such as a tube MPC-based formulation, or dynamically update tissue/laser parameters intraoperatively based on state data \cite{arnold2022identification,mayne2011tube}.

In the 3D simulation, the graph-based algorithm demonstrates the removal of large amorphous tumor regions. $96\%$ of the error shown in \autoref{fig:3dhistogram} lies within $\pm 2$mm with over $78\%$ lying within $\pm 1$mm, similar to the tremor limits of a surgeon \cite{coulson2010effect}. However, two limitations hamper real-world adoption. First, the long runtime is acceptable for pre-operational planning, but not for interoperational feedback where the plan is recalculated after every cutinclude rewriting the algorithm using PyTorch and GPU acceleration to increase speed, or to search only for small corrections to future ablations, rather than recompute the entire plan.
Another limitation is the use of the $z$-directional \acrshort{mse} evaluation metric. Calculating the objective cost of a state requires interpolating the provided objective and constraint boundaries at various $x$, $y$ points, which requires both the objective and constraint boundaries to be functions. Any ``overhangs" will have noisy interpolations, creating erroneous calculations in those regions (visible as striated regions in \autoref{fig:3dsetup}). Future work should employ nondirectional metrics such as chamfer distance or convex hull constraints.


\addtolength{\textheight}{0cm}   



\section*{APPENDIX}
\subsection{Proof of Linear Superposition in Nonangular Case}\label{apx:superposition}
At cut $i$, let $\vec{e}_i$ denote the unit vector along the laser axis and let $X_i=\left(x_i,0\right)$ denote the intersection of the laser axis with the $x$-axis. Together, $\{X_i,\vec{e}_i\}$ determine the laser center-line of cut $i$. Let $\vec{p}_i$ denote the coordinates of an arbitrary point on the tissue surface point cloud after cut $i$, and let $\vec{p}_0$ be the original point before any cuts. Let $u_i$ be all of the non-spatial input parameters to the $i^{th}$ laser cut. Equation \eqref{eq:singleablation} can then be generalized to $\Delta p_i = f(u_i,d_i)$. The coordinate of point $\vec{p}$ after $k$ cuts is
\begin{align}\label{eq:pk}
\vec{p}_k = \vec{p}_0 + \sum_{i=1}^k\left(f(u_i,d_i)\,\vec{e}_i\right).
\end{align}
Denoting the orthogonal distance between the point $\vec{p}_{i-1}$ and the laser axis $\{X_i,\vec{e}_i\}$ as ${d_i=\mathrm{dist}(\{X_i,\vec{e}_i\},\vec{p}_{i-1})}$ and substituting this into \eqref{eq:pk} produces
\begin{align}\label{eq:expandedpk}
\vec{p}_k = \vec{p}_0 + \sum_{i=1}^k(f(u_i,\mathrm{dist}(\{X_i,\vec{e}_i\},\vec{p}_{i-1}))\,\vec{e}_i).
\end{align}
The expression for $\vec{p}_{i-1}$ can be obtained from \eqref{eq:pk} as
\[
\vec{p}_{i-1} = \vec{p}_0 + \sum_{j=1}^{i-1}(\Delta p_j\vec{e}_j).
\]
Substituting this into \eqref{eq:expandedpk} gives
\[
\vec{p}_k = \vec{p}_0 + \sum_{i=1}^k(f(u_i,\mathrm{dist}(\{X_i,\vec{e}_i\},\vec{p}_0 + \sum_{j=1}^{i-1}(\Delta p_j \vec{e}_j)))\,\vec{e}_i).
\]
If all laser cuts have parallel laser axes, $\vec{e}_i = \vec{e}$ for all $i$, then this simplifies to
\[
\vec{p}_k = \vec{p}_0 + \sum_{i=1}^k(f(u_i,\mathrm{dist}(\{X_i,\vec{e}\},\vec{p}_0 + \sum_{j=1}^{i-1}(\Delta p_j \vec{e}\,)))\,\vec{e}\,).
\]
The $\mathrm{dist}\left(\cdot\right)$ term can be further simplified to
\[
\mathrm{dist}\Big(\{X_i,\vec{e}\},\vec{p}_0 + \sum_{j=1}^{i-1}(\Delta p_j \vec{e})\Big) = \mathrm{dist}\Big(\{X_i,\vec{e}\},\vec{p}_0\Big),
\]
as the summation term only displaces the point $\vec{p}$ parallel to the laser axis, $\{X_i,\vec{e}\}$, and thus does not change the orthogonal distance between the point and the axis. This results in the final expression
\[
\vec{p}_k = \vec{p}_0 + \sum_{i=1}^k(f(u_i,\mathrm{dist}(\vec{e},\vec{p}_0))\,\vec{e}\,).
\]
By additive commutativity, $\vec{p}_k$ is identical for any permutation of $i$ (order of laser cuts), shown in \autoref{fig: superposition-diagram}. 



\bibliographystyle{IEEEtran}
\bibliography{IEEEabrv, IEEEbib}

\end{document}